\documentclass[10pt,twocolumn,letterpaper]{article}
\usepackage{cvpr}  

\def\cvprPaperID{7169} 
\def\confName{CVPR}
\def\confYear{2022}
\usepackage{times}
\usepackage{epsfig}
\usepackage{graphicx}
\usepackage{amsmath}
\usepackage{amssymb}
\usepackage{url}            
\usepackage{booktabs}       
\usepackage{amsfonts}       
\usepackage{nicefrac}       
\usepackage[T1]{fontenc}
\usepackage{microtype}      
\usepackage{times}
\usepackage{epsfig}
\usepackage{graphicx}
\usepackage{amsmath}
\usepackage{amssymb}
\usepackage{caption}
\usepackage{subcaption}
\usepackage{wrapfig}
\usepackage{lipsum}
\usepackage{multirow}
\usepackage{booktabs}
\usepackage{tabu}
\usepackage{tabularx}
\usepackage{spreadtab}

\usepackage[pagebackref,breaklinks,colorlinks]{hyperref}

\usepackage[capitalize]{cleveref}
\crefname{section}{Sec.}{Secs.}
\Crefname{section}{Section}{Sections}
\Crefname{table}{Table}{Tables}
\crefname{table}{Tab.}{Tabs.}

\begin{document}

\title{UKPGAN: A General Self-Supervised Keypoint Detector}

\author{Yang You, Wenhai Liu, Yanjie Ze, Yong-Lu Li, Weiming Wang\thanks{Cewu Lu and Weiming Wang are the corresponding authors. Cewu Lu is member of Qing Yuan Research Institute and MoE Key Lab of Artificial Intelligence, AI Institute, Shanghai Jiao Tong University, China and Shanghai Qi Zhi institute.},\, Cewu Lu\footnotemark[1]  \\ 
Shanghai Jiao Tong University, China\\ 
\{qq456cvb, sjtu-wenhai, zeyanjie, yonglu\_li, wangweiming, lucewu\}@sjtu.edu.cn 
}

\maketitle

\begin{abstract}
Keypoint detection is an essential component for the object registration and alignment. 
In this work, we reckon keypoint detection as information compression, and force the model to distill out irrelevant points of an object. Based on this, we propose UKPGAN, a general \textbf{self-supervised} 3D keypoint detector where keypoints are detected so that they could reconstruct the original object shape. Two modules: \textbf{GAN-based keypoint sparsity control} and \textbf{salient information distillation} modules are proposed to locate those important keypoints. Extensive experiments show that our keypoints align well with human annotated keypoint labels, and can be applied to SMPL human bodies under various non-rigid deformations. Furthermore, our keypoint detector trained on clean object collections generalizes well to real-world scenarios, thus further improves geometric registration when combined with off-the-shelf point descriptors. Repeatability experiments show that our model is stable under both rigid and non-rigid transformations, with local reference frame estimation. 
Our code is available on \href{https://github.com/qq456cvb/UKPGAN}{https://github.com/qq456cvb/UKPGAN}.
\end{abstract}

\section{Introduction}
Recently, 3D object analysis and scene understanding receive more and more attentions. Though plenty of methods~\cite{dai2017scannet,li2018so,murez2020atlas,he2019geonet} on object analysis have been proposed, there is still a lack of capability of processing and understanding objects, especially under an unsupervised setting.

3D keypoints, unlike part annotations, provide a sparse but meaningful representations of an object. They are widely leveraged in many tasks such as object matching, object tracking, shape retrieval and registration~\cite{mian2006three, bueno2016detection, wang2018learning}. Keypoint detections have its origin in 2D image processing~\cite{rublee2011orb,lowe2004distinctive,harris1988combined}. In 3D domain, traditional methods like Harris-3D~\cite{sipiran2011harris}, HKS~\cite{sun2009concise}, Salient Points~\cite{castellani2008sparse}, Mesh Saliency~\cite{lee2005mesh}, ISS~\cite{zhong2009intrinsic}, Sift-3D~\cite{rister2017volumetric} and Scale Dependent Corners~\cite{novatnack2007scale} propose to detect keypoints based on geometric variations. However, these hand-crafted detectors rely heavily on hard-coded parameters and their performance is not comparable to current learning-based methods. 

\begin{figure}[ht]
    \centering
    \includegraphics[width=\linewidth]{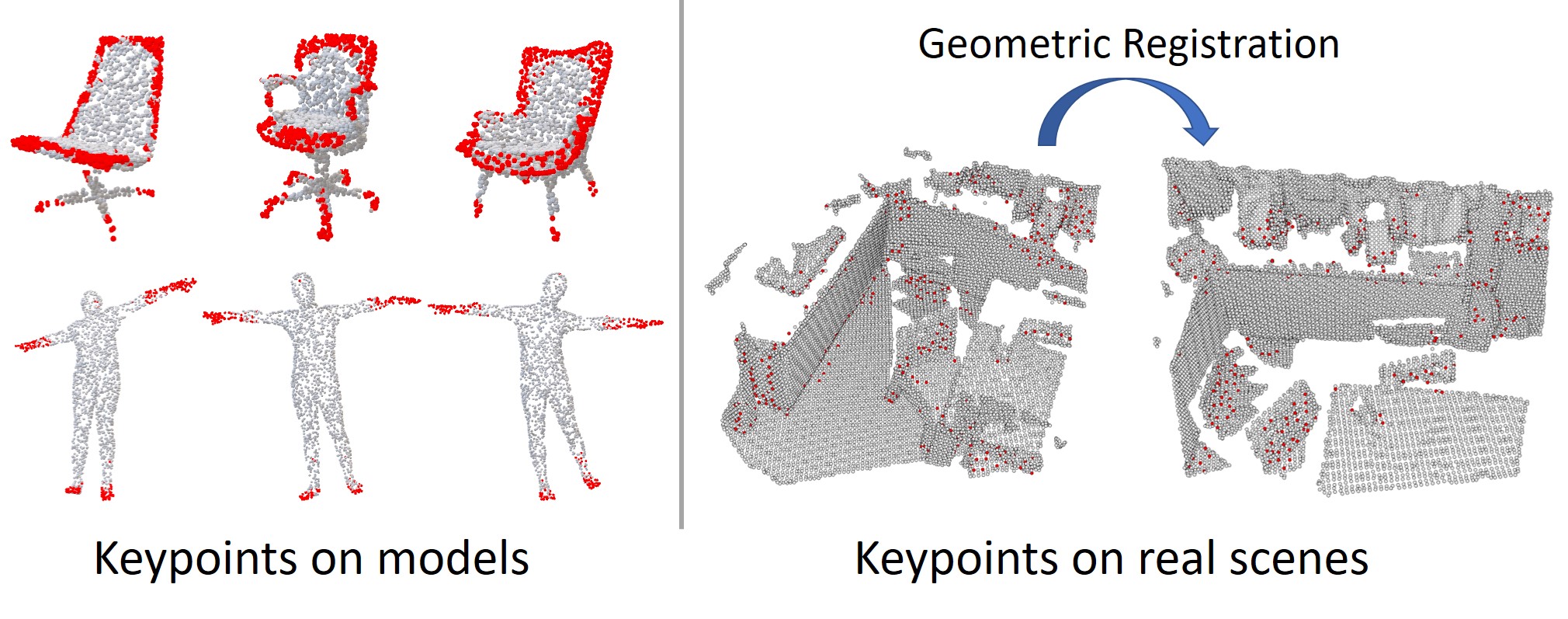}
    \caption{Our model outputs unsupervised keypoints and embeddings given a point cloud, in either rigid or non-rigid deformations. Left are keypoint predictions on clean models (indicated by red) and right are keypoint predictions on real scenes, best viewed in color. These keypoints are \textit{consistent} and could be used for registration.}
    \label{fig:intro}
\end{figure}

Recently, some learning-based methods like USIP~\cite{li2019usip} and D3Feat~\cite{bai2020d3feat} have been proposed. USIP regresses keypoint locations from pre-segmented local groups and then utilizes a probabilistic chamfer loss. However, their method requires the farthest point sampling and may output points that are not on the input. D3Feat instead gives saliency scores and descriptors densely for each point. Both USIP and D3Feat predict 3D keypoints by solving the auxiliary task of correctly estimating rotations in a Siamese architecture. They both require real-world point clouds during training, and do not have much control on the output keypoints.

To solve these problems, we follow a totally different route to obtain 3D keypoints, which is named \textbf{Unsupervised Key Point GANeration (UKPGAN)}. A keypoint saliency distribution is given through a detector network, with a novel adversarial \textbf{GAN loss to control its sparsity}. Then, to make these keypoints informative, we leverage a \textbf{salient information distillation} process to reconstruct the original point cloud from these sparse keypoints, forming an encoder-decoder architecture. Our model can be seen as an \textit{information compression scheme}, keeping most information of the object with the least keypoints. The rationale behind our method is simple but powerful: one should be able to fully recover an object's structure from a small set of keypoints. This also coincides with that mentioned in \cite{wolff2019information}: ``\textit{much of human learning, perception, and cognition, may be understood as information compression}''. Results show that our model could output stable informative keypoints from unseen objects, and generalize well to real-world scenarios (Figure~\ref{fig:intro}).

Compared to previous methods, UKPGAN has the following advantages: 1) our detector is proven to be rotation invariant without any data augmentations, by first estimating a Local Reference Frame (LRF), which also makes our local keypoint representation disentangled from rotations; 2) detected keypoints are intra-class consistent and stable on both rigid and non-rigid objects, with high repeatability; 
3) our model trained on clean object collections (i.e. ModelNet) generalizes well to real-world point clouds, free from the usage of real-world training data.

We first evaluate our method on ShapeNet models with keypoint labels. Our model achieves remarkable results in keep consistent with human labeled part and keypoints. UKPGAN cannot only be applied to rigid but non-rigid objects by keeping consistency on SMPL human body deformable meshes. As an application of our model, we also evaluate UKPGAN on 3DMatch and ETH datasets, which are real-world geometric registration benchmarks. Experiments show that when trained on clean objects (i.e. ModelNet), our model generalizes well to real-world scenarios, and further improves the registration performance of current state-of-the-art methods. At last, extensive experiments are conducted to demonstrate that UKPGAN achieves high rotation repeatability, which is an important and desired property of keypoints.
 
\section{Related Work}
\subsection{Hand-crafted Keypoint Detectors}
Prior to  deep learning, researchers proposed numerous methods to detect stable interest points on objects, in both 2D and 3D domains. SIFT~\cite{lowe2004distinctive}, ORB~\cite{rublee2011orb} and SURF~\cite{bay2006surf} extract features by detecting local pattern variations on 2D images. They are robust to scale and rotation changes and give consistent keypoints on two identity objects.  3D Harris~\cite{sipiran2011harris} extends Harris corner detector to 3D meshes. HKS~\cite{sun2009concise} proposes a novel point signature based on the properties of the heat diffusion process on a shape. Salient Points~\cite{castellani2008sparse} model interest points by a Hidden Markov Model (HMM), which is trained in an unsupervised way by using contextual 3D neighborhood information. Mesh Saliency~\cite{lee2005mesh} defines mesh saliency in a scale-dependent manner using a center-surround operator on Gaussian-weighted mean curvatures. CGF~\cite{khoury2017learning} learns to represent the local geometry around a point in an unstructured point cloud. 3D SIFT~\cite{rister2017volumetric} is an analogue of the scale-invariant feature transform (SIFT) for three-dimensional images. ISS~\cite{zhong2009intrinsic} introduces intrinsic shape signature, which uses a view-independent representation of the 3D shape to match shape patches from different views directly. However, these methods only consider the local geometric information without semantic knowledge, leading to a discrepancy from human perceptions.

\subsection{Learning-based Keypoint Detectors}
Recently, some deep learning based detectors have been proposed to bypass hand-crafted keypoint detection rules, in both 2D and 3D domains. On 2D images, some unsupervised keypoint detection methods are proposed. Jakab et al.~\cite{jakab2018unsupervised} extracts semantically meaningful keypoints by passing a target image through a tight bottleneck to distill the geometry of the object. Zhang et al.~\cite{zhang2018unsupervised} uses an auto-encoding module with channel-wise softmax operation to discover landmarks. Suwajanakorn et al.~\cite{suwajanakorn2018discovery} discover latent 3D keypoints from 2D images by enforcing multi-view consistency. Georgakis~et al.\cite{georgakis2018end} employ a Siamese architecture augmented by a sampling layer and a novel score loss function to detect keypoints on depth maps. In 3D domain, methods like SyncSpecCNN~\cite{yi2017syncspeccnn} and deep functional dictionaries~\cite{sung2018deep} rely on ground-truth keypoint supervision. For unsupervised methods, USIP~\cite{li2019usip} regresses keypoint locations from pre-segmented local groups and then utilizes a probabilistic Chamfer loss. D3Feat~\cite{bai2020d3feat} instead gives saliency scores and descriptors densely for each point. It relies on an auxiliary task of correctly estimating rotations in a Siamese architecture, ignoring semantic information.
There exists another line of search~\cite{chen2020unsupervised,fernandez2020unsupervised,jakab2021keypointdeformer} that outputs a predefined fixed number of keypoints by regressing the absolute coordinates. However, these methods are not robust to rigid transformations and fail to generalize to real-world scenarios. 

\section{Approach}

\begin{figure*}[!h]
    \centering
    \includegraphics[width=0.85\linewidth]{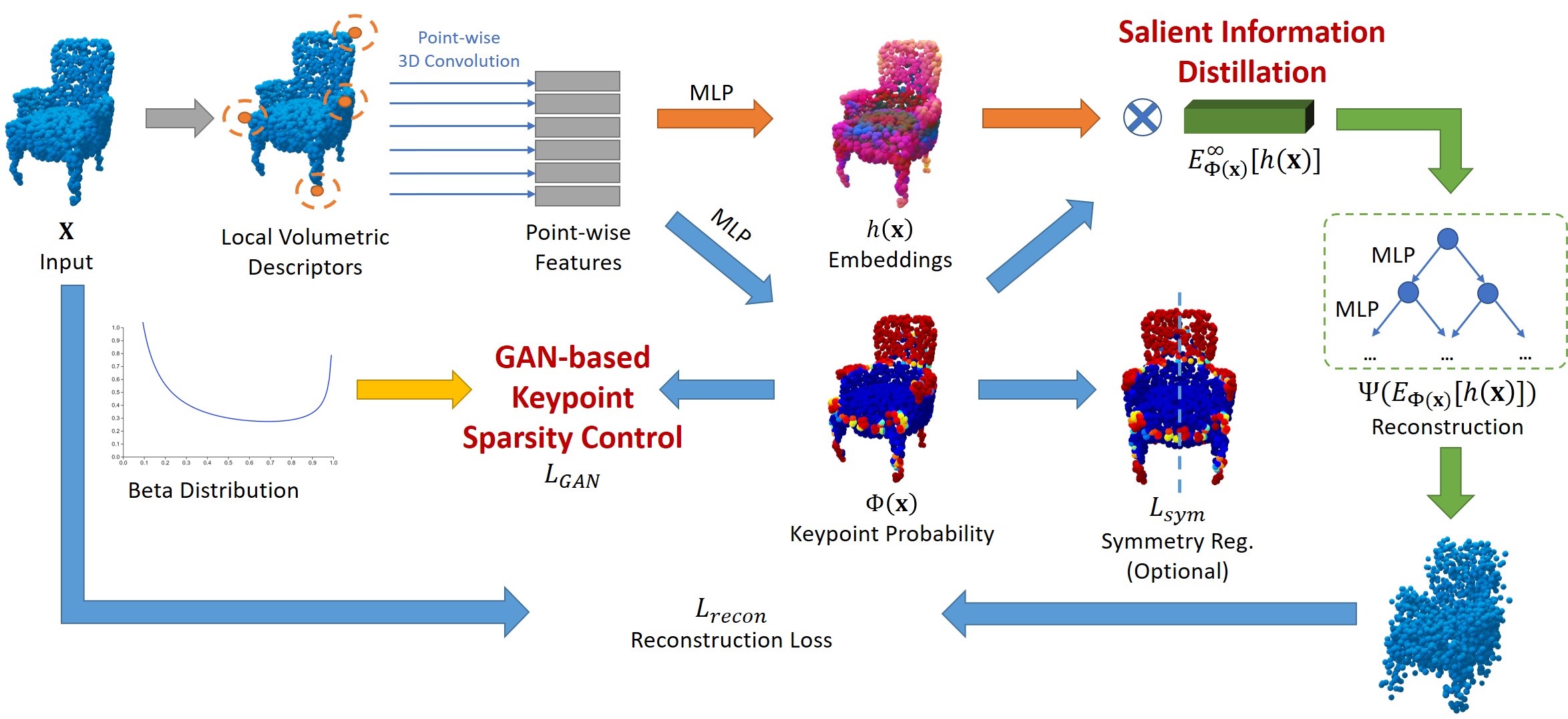}
    \caption{\textbf{Our whole pipeline on keypoint and embedding generations.} We first extract point-wise rotation invariant features and then output keypoint probabilities and semantic embeddings through two MLPs, respectively. GAN is leveraged to control keypoint sparsity, and salient information distillation is utilized to distill most salient features. A decoder is concatenated to reconstruct original point clouds.}
    \label{fig:pipeline}
\end{figure*}
\subsection{Overview}
Given a point set $\mathbf{X} = \{\mathbf{x}_n | \mathbf{x}_n\in \mathbb{R}^3, n = 1, 2,\dots N\}$ with $\mathbf{x}$ sampled from some manifold $\mathcal{M}$, we seek a keypoint set $\Tilde{\mathbf{X}}\subseteq\mathbf{X}$, where $|\Tilde{\mathbf{X}}|$ is the number of required keypoints.

Here, we propose an unsupervised encoder-decoder architecture. In the encoder, which is also the detector, a keypoint probability $s$ is predicted for each point. To keep the detected keypoints sparse, GAN-based keypoint sparsity control is leveraged. In the decoder, also a reconstruction network, we utilize salient information distillation to reconstruct the original point cloud, in an unsupervised way. The intuition is that, a set of good keypoints should contribute to the unique information of an object, making reconstruction possible. The overview of our method is shown in Figure~\ref{fig:pipeline}.


\subsection{Rotation Invariant Feature Extraction}

In order to be robust under rigid transformations, we first generate a Local Reference Frame (LRF) by co-variance eigen-decomposition on each point $\mathbf{x}$'s spherical neighborhoods $\mathcal{S} = \{\mathbf{x}_i : \|\mathbf{x}_i - \mathbf{x}\|_2 \leq r\}$. Then points in the local neighborhood $\mathbf{x}_i \in \mathcal{S}$ are transformed to their canonical position $\mathbf{x}_i'$ according to the estimated LRFs.
Next, we follow the same strategy as that in PerfectMatch~\cite{gojcic2019perfect} to discretize these points in a Smoothed Density Value (SDV) grid, centered on the point $\mathbf{x}$ and aligned with the LRF. The voxelization is based on Gaussian smoothing kernel. Afterwards, we would get a voxelized descriptor $\mathcal{F}(\mathbf{x})\in \mathbb{R}^{W\times H\times D}$ for each point $\mathbf{x}$. For mode details, we refer the reader to PerfectMatch~\cite{gojcic2019perfect}. These point-wise 3D descriptors are batched together and fed through 3D convolution layers to be further refined. Thanks to the estimated LRFs, this step provides \textit{local} \textit{rotation-invariant} features, which are critical for rotation repeatability.

\subsection{Dual Branches on Estimating Probabilities and Embeddings}

After extracting rotation invariant point-wise features, we use dual Multi-Layer Perceptron (MLP) networks to estimate a keypoint salient probability $\Phi(\mathbf{x}) \in [0, 1]$ and a high-dimensional embedding $h(\mathbf{x}) \in \mathbb{R}^F$, which will be used for reconstruction.

\paragraph{Sparsity on $\Phi(\mathbf{x})$.}
In order to compress the entire point cloud with a minimum set of keypoints, $\Phi(\mathbf{x})$ needs to be sparse. What is a good way to make $\Phi(\mathbf{x})$ sparse? One would consider L1 regularization. However, it tends to output more probabilities around zero and does not have much control over non-zero probabilities. In order to output distinguishable keypoints and suppress those meaningless points, we would like $\Phi(\mathbf{x})$ to accumulate around both 0's and 1's. A straight-forward solution is to define a controllable keypoint distribution that accumulates around 0's and 1's, then force the network prediction to match this prior. Inspired by~\cite{zamorski2020adversarial}, we take Beta distribution (shown in Figure~\ref{fig:pipeline}) as our keypoint distribution prior. In Beta distribution, there are two parameters $\alpha$ and $\beta$, which control the accumulation of positive (1) and negative (0) samples, respectively. For more details of controllability provided by Beta distribution, please refer to our supplementary.

\paragraph{GAN-based Keypoint Sparsity Control}
A direct solution to force the sparsity is to compute the KL divergence between the predicted keypoint distribution and the Beta prior. However, since we are predicting keypoint \textit{samples} instead of distribution \textit{parameters}, the closed form of KL divergence between Beta prior and $\Phi(\mathbf{x})$ does not exist. We resort to adversarial loss to resolve this. 

GAN~\cite{goodfellow2014generative} is leveraged to generate fake keypoint distributions that look real to our Beta prior (i.e., $p(\mathbf{x})$). It requires a discriminator network $D$ and a generator network (i.e., $\Phi(\cdot)$). Notice that in our adversarial training settings, each sample is a keypoint distribution on a point cloud, which is itself sampled from a repository. The input to the discriminator network $D$ is the whole keypoint distribution set on a single point cloud. The reason not to input the single keypoint is that we want each object's keypoint distribution to follow our Beta prior, but not the distribution of the points from all objects. 


In practice, we employ WGAN-GP~\cite{arjovsky2017wasserstein} instead of the naive GAN loss as it is more robust. The loss follows:
\begin{align}
    L_{GAN} =&\ \min_\Phi\max_D(\mathbb{E}_{\mathcal{M}}[D(\{p(\mathbf{x})|\mathbf{x}\in\mathcal{M}\})]\\ & -\mathbb{E}_{\mathcal{M}}[D(\{\Phi(\mathbf{x})|\mathbf{x}\in\mathcal{M}\})]
     + \lambda (\|\nabla D\|_2 - 1)^2),
\end{align}
which penalizes the gradient of the discriminator.

\subsection{Reconstruction Network}
Given a keypoint distribution $\{\Phi(\mathbf{x})\in \mathbb{R} | \mathbf{x}\sim\mathcal{M}\}$ and high-dimensional embeddings  $\{h(\mathbf{x})\in \mathbb{R}^F | \mathbf{x}\sim\mathcal{M}\}$, a point cloud decoder is introduced to reconstruct the original shape. Denoting the point cloud decoder as $\Psi: \mathbb{R}^N\times\mathbb{R}^{N\times F}\rightarrow \mathbb{R}^{N\times 3}$, the reconstruction loss can be expressed as follows:
\begin{align}
\label{eq:rec}
    L_{recon} &= CD(\Psi(\{\Phi(\mathbf{x})|\mathbf{x}\sim\mathcal{M}\}, \{h(\mathbf{x})|\mathbf{x}\sim\mathcal{M}\}), \mathbf{X}),
\end{align}
where $CD$ is the Chamfer distance.


\paragraph{Salient Information Distillation}
In Equation~\ref{eq:rec}, $\Psi$ takes both keypoint distribution and high-dimension embeddings as the input.
Recall that our goal is to find a sparse set of salient keypoints that possibly reconstruct the original shape. To fulfill this, we get some inspiration from the \textit{max} operation in PointNet~\cite{qi2017pointnet} and propose a \textit{salient information distillation} module. This module is leveraged to force the network to give both probable (large $\Phi(\mathbf{x})$) and semantic-rich (large $h(\mathbf{x})$) keypoints.


We define $\Psi$ as:
\begin{align}
    \Psi = \text{TopNet}(\max_{\mathbf{x}\sim\mathcal{M}}[\Phi(\mathbf{x})\cdot h(\mathbf{x})]),
\end{align}
where we slightly abuse the notation such that $\Phi(\mathbf{x})$ is broadcasted when multiplying with $h(\mathbf{x})$. The max operation is also conducted channel-wisely, so that $\max_{\mathbf{x}\sim\mathcal{M}}[\Phi(\mathbf{x})\cdot h(\mathbf{x}) \in \mathbb{R}^F$. TopNet represents the point decoder structure similar to that of  Tchapmi \textit{et al.}~\cite{tchapmi2019topnet}.

 In addition, for semantic $h(x)$, we care about the absolute value of $h(\mathbf{x})$ (features with large negative magnitude should not be suppressed) and the final decoder is 

\begin{align}
     \Psi = &\text{TopNet}(\max_{\mathbf{x}\sim\mathcal{M}}[\Phi(\mathbf{x})\cdot \max(h(\mathbf{x}), 0)] \\
    &\oplus \max_{\mathbf{x}\sim\mathcal{M}}[\Phi(\mathbf{x})\cdot \max(-h(\mathbf{x}), 0)]),
\end{align}
where $\oplus$ means concatenation.

 Intuitively, our decoder forces the network to mark those semantic-rich (large $h(x)$) as salient keypoints (large $\Phi(x)$), otherwise the product will be small and get suppressed because of the \textit{max} operation. On the other side, indistinguishable points with similar local context are therefore discarded. For example, a rectangle can be perfectly reconstructed given the four corners. Points between the corners provide little information about the overall shape. Detailed analysis on salient information distillation will be given in Section~\ref{sec:ablation}.

\subsection{Symmetric Regularization}
\label{sec:sym}




Although we first extract rotation invariant local descriptors from original point clouds, it is not symmetric invariant. For most common objects, we have a strong prior such that detected keypoints and features should be symmetric, leading to the following loss:
\begin{align}
    L_{sym} = \frac{1}{|\mathbf{S}|}\sum_{(\mathbf{x}, \mathbf{x'})\in\mathbf{S}}(\|\Phi(\mathbf{x}) - \Phi(\mathbf{x'})\| +\|h(\mathbf{x})-h(\mathbf{x'})\|_1),
\end{align}
where $\mathbf{S}$ is the set of all symmetric point pairs.
Note that symmetric regularization is only used for training; in testing, symmetric information about objects is not required.

The final loss is an empirical sum of three terms:
\begin{align}
    L = \eta_1\cdot L_{recon} + \eta_2\cdot L_{GAN} + \eta_3\cdot L_{sym}.
\end{align}




\subsection{Implementation Details}
\paragraph{Network Architecture}
Our model takes a point cloud $\mathbf{X}\in\mathbb{R}^{N\times 3}$ as input where $N=2048$. Then  a voxelized descriptor is extracted for each point with $\{\mathcal{F}(\mathbf{x_n})\}_{n=1}^N\in \mathbb{R}^{N\times  W\times H\times D}$. Then these descriptors are fed into seven 3D convolution layers with channels  32, 32, 64, 64, 128, 128, 128. To predict $\Phi(\mathbf{x})$, three-layer MLP with channels 512, 256, 1 is employed; for $h(\mathbf{x})$, three-layer MLP with channels 512, 256, 128 is employed, and the embedding dimension is 128. These two branches share the first two layers. 

For WGAN-GP network, we use five \textit{conv1d} layers (with channels 512, 256, 128, 64, 1) and a \textit{max-pooling} layer for the critic function $D$. The gradient penalty coefficient $\lambda = 1$.

For the decoder, we leverage a similar structure with TopNet~\cite{tchapmi2019topnet}. Specifically, the decoder has 6 levels and each MLP in the decoder tree generates a small
node feature embedding of size 8. When generating $N = 2048$ points, the root node has 4 children and all other internal nodes in subsequent level generate 8 children. Each MLP in the decoder is a has 3 stages with 256, 64, and 8 channels respectively.

\paragraph{Hyperparameters and Training}
For ShapeNet models, we choose $\eta_1=10.$, $\eta_2=1$, $\eta_3=0.1$ through the validation set; for SMPL human body dataset,  we choose $\eta_1=10.$, $\eta_2=1$, $\eta_3=0$. In all our experiments without specification, the Beta prior distribution is fixed with $\alpha=0.01$ and $\beta=0.05$. The parameters of the network are optimized using Adam~\cite{DBLP:journals/corr/KingmaB14}, with learning rate 1e-4. 

\section{Experiments}
In experiments, we evaluate our method on various datasets to show its capability on detecting stable interest points on general objects and real-world scenes. We first compare our keypoints with human annotated ones on ShapeNet models to show the intra-class semantic consistency between the keypoints. Next, we show that UKPGAN can be used to track stable interest points on non-rigid human bodies under different poses from SMPL models~\cite{SMPL:2015}. Then, we demonstrate our model's performance on 3DMatch and ETH registration benchmarks. Finally, keypoint repeatability under arbitrary rotations is evaluated, and an ablation study follows to validate our proposed components.

\subsection{Comparison with Human Annotated Keypoints}
\label{sec:human}
In this section, we compare detected keypoints with those human annotated ones, in order to see if there is any semantic correspondence among keypoints.
\paragraph{Dataset}
 Two datasets are utilized: ShapeNet-chair keypoint and KeypointNet~\cite{you2020keypointnet} dataset. ShapeNet-chair keypoint set is proposed by SyncSpecCNN~\cite{yi2017syncspeccnn}, which consists of thousands of keypoints annotated on ShapeNet chairs by experts. KeypointNet annotates millions of keypoints on models from 16 object categories of ShapeNet. We evaluate on airplanes, chairs and tables for KeypointNet and on chairs for ShapeNet-chair keypoint dataset. For both datasets, we randomly split them into 75\%, 10\% and 15\% for train, val and test.
\paragraph{Metric}
We evaluate the performance by mean Intersection over Union (mIoU). Intersection is counted if the geodesic distance of a detected keypoint to its closest ground-truth is less than some geodesic threshold. Union is simply the union of detected and ground-truth keypoints. 

\begin{figure*}
    \centering
    \includegraphics[width=0.8\linewidth]{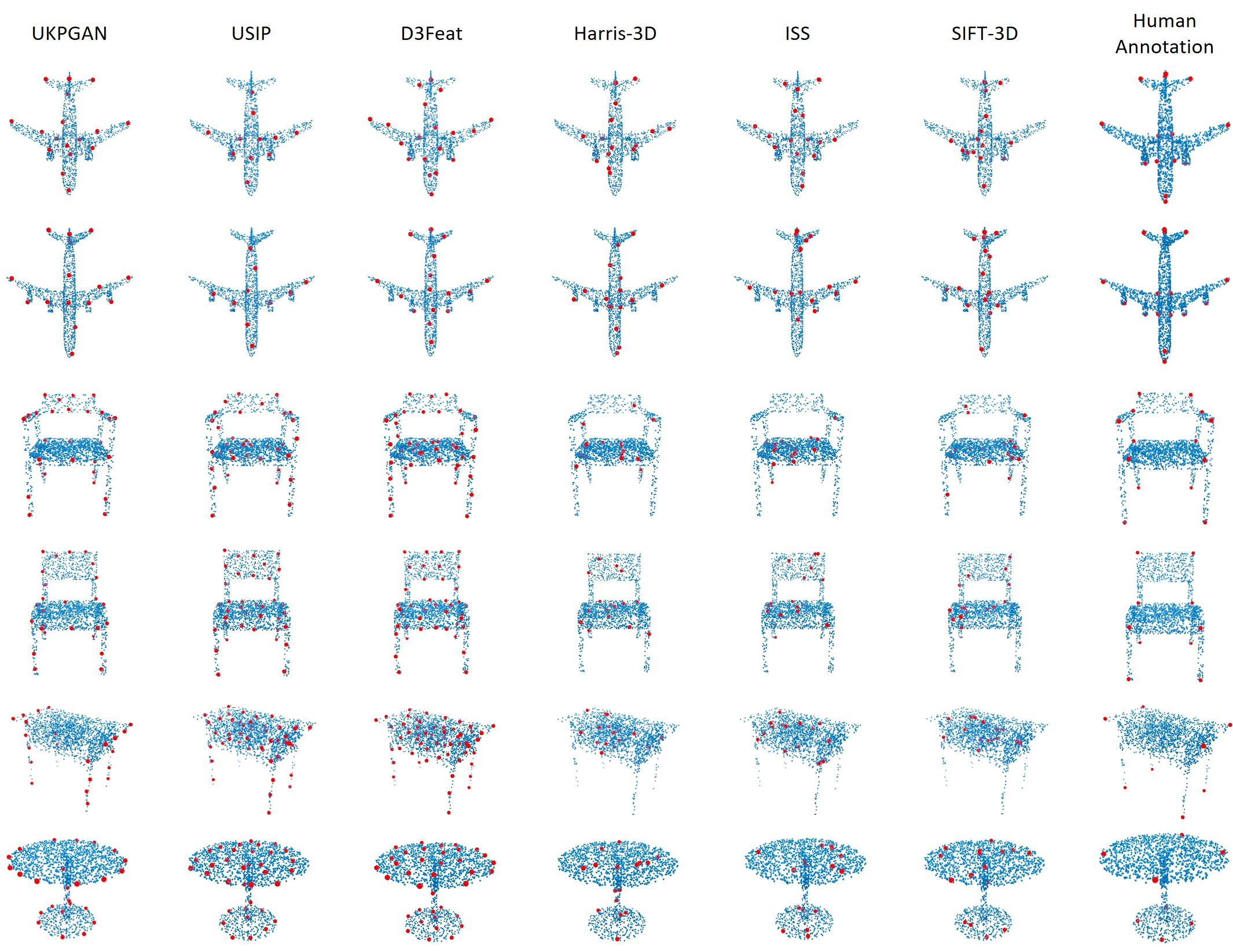}
    \caption{\textbf{Visualizations of six algorithms on unsupervised keypoint detection on ShapeNet models.}}
    \label{fig:vis}
\end{figure*}
\begin{figure*}
\centering
\begin{minipage}{.67\textwidth}
  \centering
        \includegraphics[width=.43\linewidth]{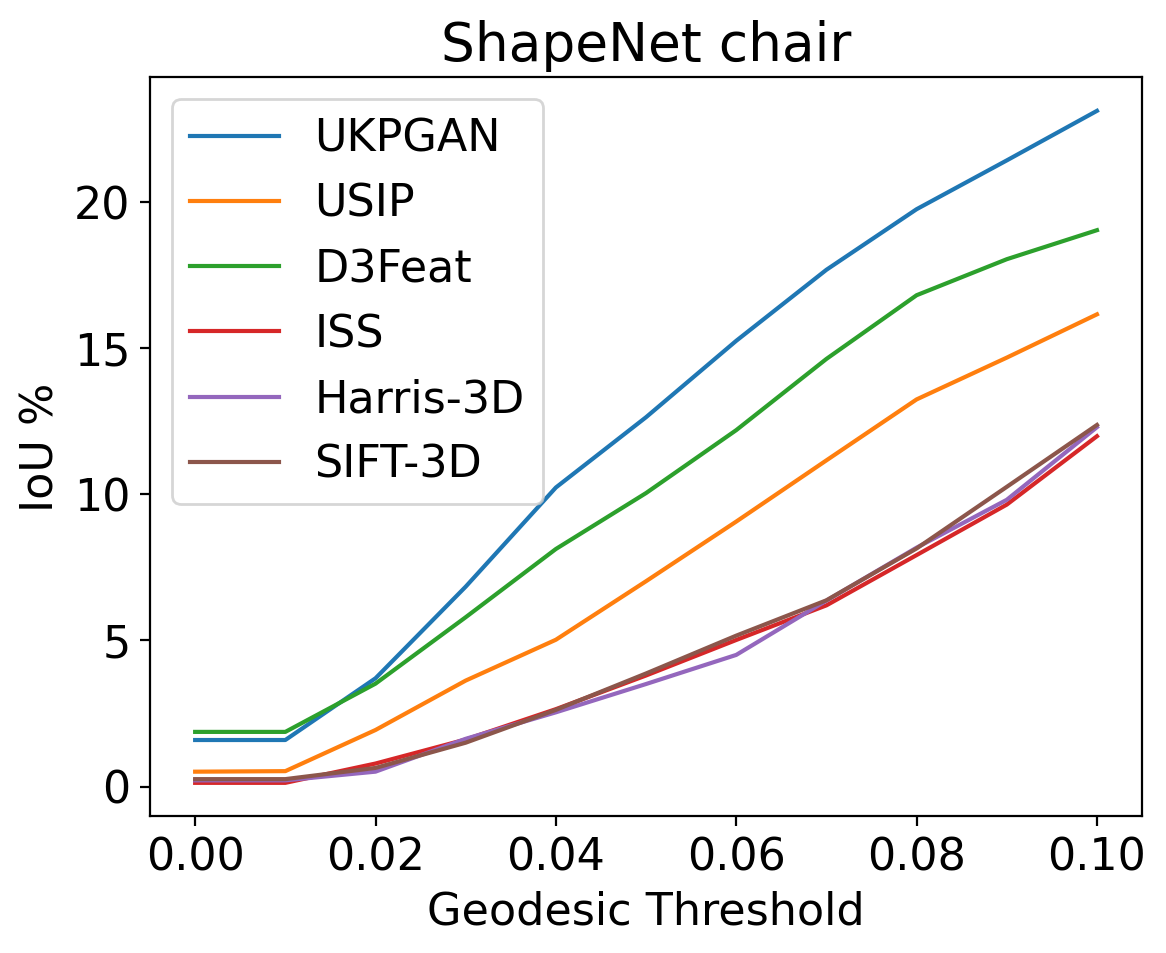}%
        \includegraphics[width=.43\linewidth]{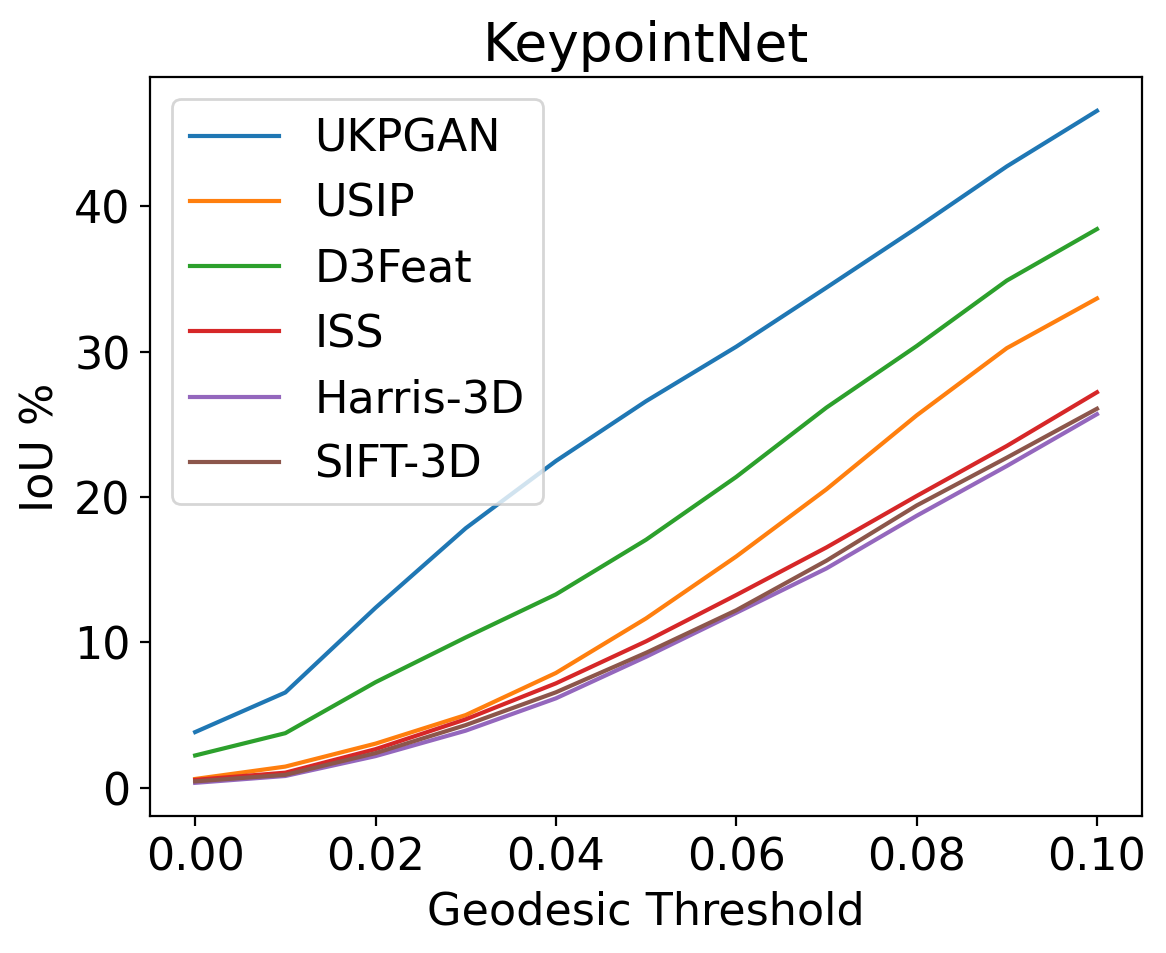}
  \captionof{figure}{\textbf{mIoU results on ShapeNet chair dataset and KeypointNet.}}
    \label{fig:iou}
\end{minipage}%
\begin{minipage}{.33\textwidth}
  \centering
  \includegraphics[width=.9\linewidth]{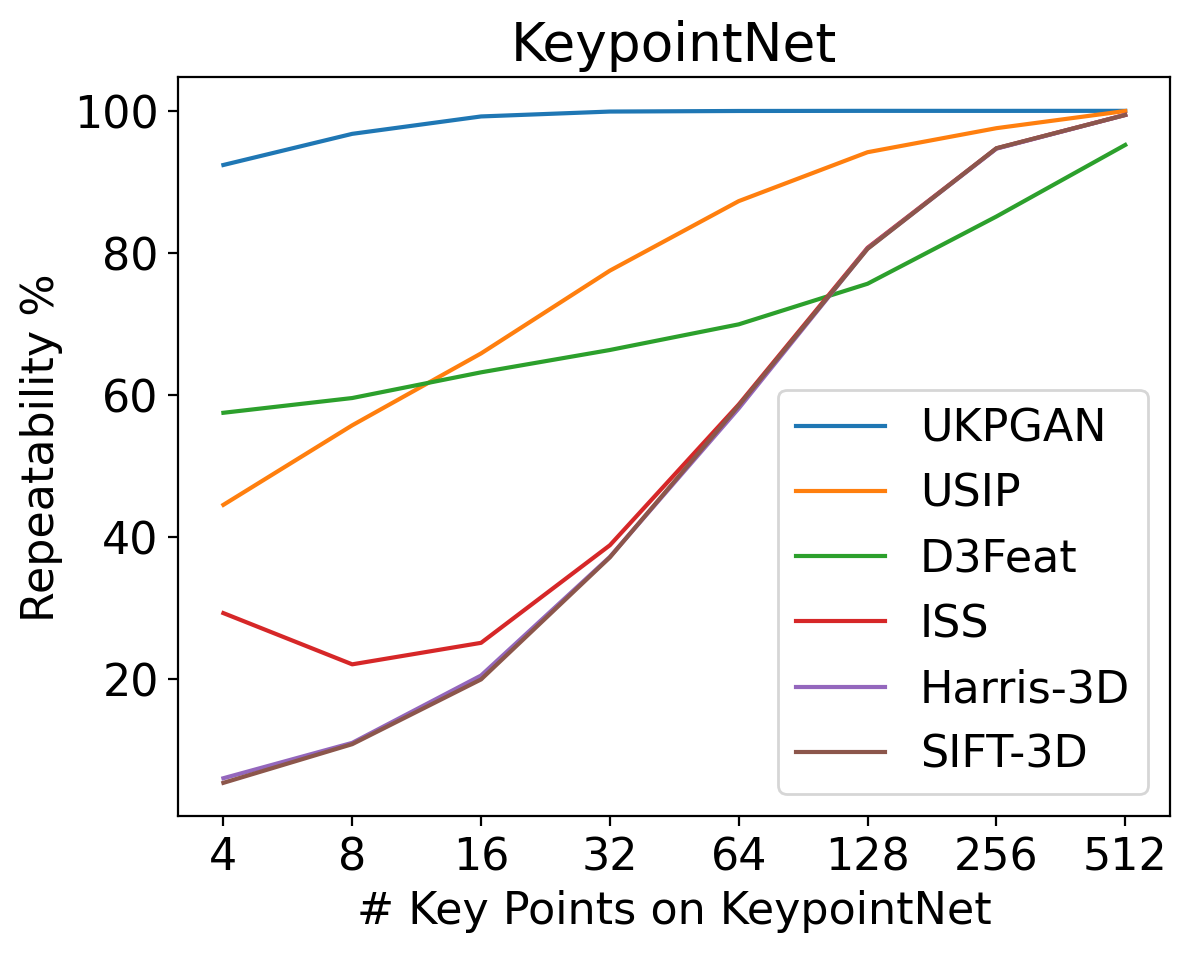}
  \captionof{figure}{\textbf{Rotation repeatability.}}
  \label{fig:rot_rep}
\end{minipage}
\end{figure*}

\paragraph{Evaluation and Results}
We compare UKPGAN with USIP~\cite{li2019usip}, D3Feat~\cite{bai2020d3feat}, Harris-3D~\cite{sipiran2011harris}, ISS~\cite{zhong2009intrinsic} and SIFT-3D~\cite{rister2017volumetric}. Training is done independently for each category. UKPGAN, USIP and D3Feat output keypoint probabilities which are refined through Non-Maximum-Suppression (NMS) with radius 0.1 and threshold ($p=0.5$). 
Quantitative results are given in Figure~\ref{fig:iou}. We see that UKPGAN aligns better with human annotated keypoints. It achieves much higher IoU than other methods. Qualitative visualizations are shown in Figure~\ref{fig:vis}. UKPGAN gives keypoints that are intra-class consistent and edge/corner salient. 

\subsection{Detecting Stable Keypoints with Semantics under Different Human Poses}

\paragraph{Dataset} Skinned Multi-Person Linear model (SMPL) is a skinned vertex-based model that accurately represents a wide variety of body shapes in natural human poses. Human poses are controlled with three parameters, and we generate training data on the fly by changing these parameters. 2048 points are sampled uniformly from the original mesh. 

\paragraph{Metric}
SMPL provides point-to-point correspondence across different human models. Given a pair of models (Model A, B), we evaluate detectors' stability and consistency by calculating Intersection of Union (IoU). An intersection is counted if a detected keypoint in Model A has its corresponding point in Model B detected too. The union is the summation of all the detected keypoints in both models. In order to take noise into account, consistency loss is also evaluated. It is calculated as the average distance between a detected point in Model A and its nearest detected neighbor in Model B, under ground-truth correspondences.

\paragraph{Evaluation and Results}
We evaluate the performance of UKPGAN method and compare it with USIP~\cite{li2019usip}, D3Feat~\cite{bai2020d3feat}, Harris-3D~\cite{sipiran2011harris}, ISS~\cite{zhong2009intrinsic} and SIFT-3D~\cite{rister2017volumetric}. Keypoints are selected with threshold $p = 0.5$ with no NMS applied. Additionally, we adapt the number of predicted keypoints of baselines so that they are directly comparable to our model. More comparisons on fixed number of keypoints (10, 20, 40) with NMS enabled are given in our supplementary.
Quantitative results are given in Table~\ref{tab:smpl}. Our method achieves the best IoU and consistency loss, suggesting it is robust and stable.

\begin{figure}[ht]
    \centering
    \includegraphics[width=0.8\linewidth]{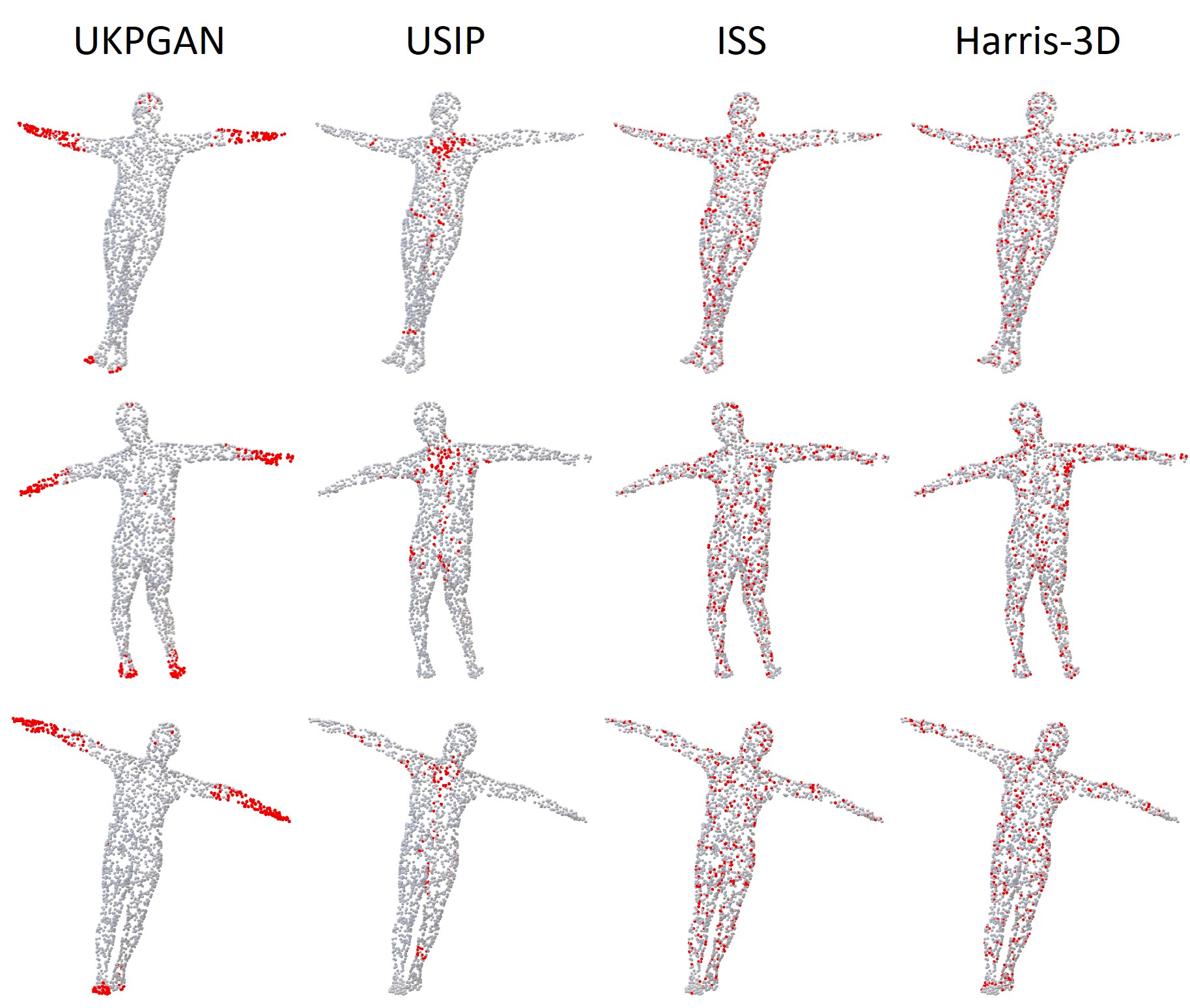}
    \caption{\textbf{Keypoint prediction results on SMPL dataset.}}
    \label{fig:smpl}
\end{figure}

Qualitative results are shown in Figure~\ref{fig:smpl}. We see that UKPGAN is able to generate corresponding stable interest points under different human poses. In contrast, though USIP gives relatively consistent keypoints with 20+ IoU, these keypoints lack semantic interpretation. Traditional keypoint detection methods like ISS and Harris are unaware of semantics and fail to give consistent keypoints.

\begin{table}[ht]
\centering
\resizebox{0.9\linewidth}{!}{
\begin{tabular}{l|c|c}
    \hline
     & IoU (\%) $\uparrow$ & Consistency Loss ($\times 1e^{-3}$) $\downarrow$\\
    \hline
    USIP~\cite{li2019usip} & 23.9 & 4.6\\
    D3Feat~\cite{bai2020d3feat} & 20.3 & 3.8 \\
    Harris-3D~\cite{sipiran2011harris} & 8.1 & 3.2  \\
    ISS~\cite{zhong2009intrinsic} & 8.1 & 3.3 \\
    SIFT-3D~\cite{rister2017volumetric} & 8.2 & 3.3\\
    \hline
    Ours & \textbf{66.6} & \textbf{1.2}\\
    \hline
\end{tabular}}
\caption{\textbf{IoU (\%) and Consistency Loss ($\times 1e^{-3}$) results for SMPL dataset.} Our keypoint detector is more stable under different deformations.}
\label{tab:smpl}
\end{table}

\begin{figure*}
    \centering
    \includegraphics[width=0.9\linewidth]{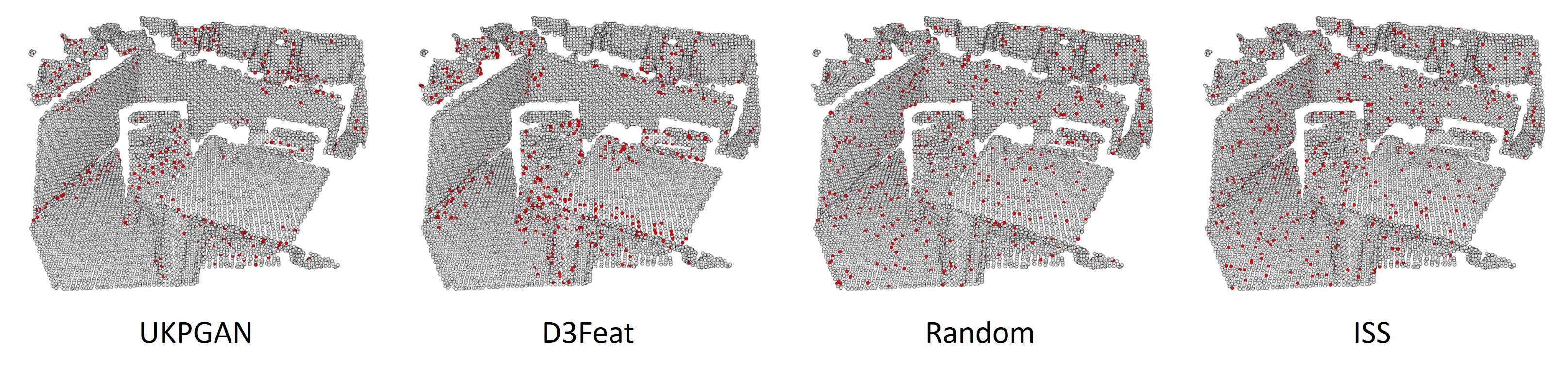}
    \caption{\textbf{Keypoint detection comparisons on 3DMatch Dataset.} Notice that our method is trained on synthetic models only, and achieve competitive results with that trained on real scenes (i.e., D3Feat). UKPGAN is able to give distinguishable keypoints for registration.}
    \label{fig:regis}
\end{figure*}

\begin{figure*}[ht]
    \centering
    \includegraphics[width=0.9\linewidth]{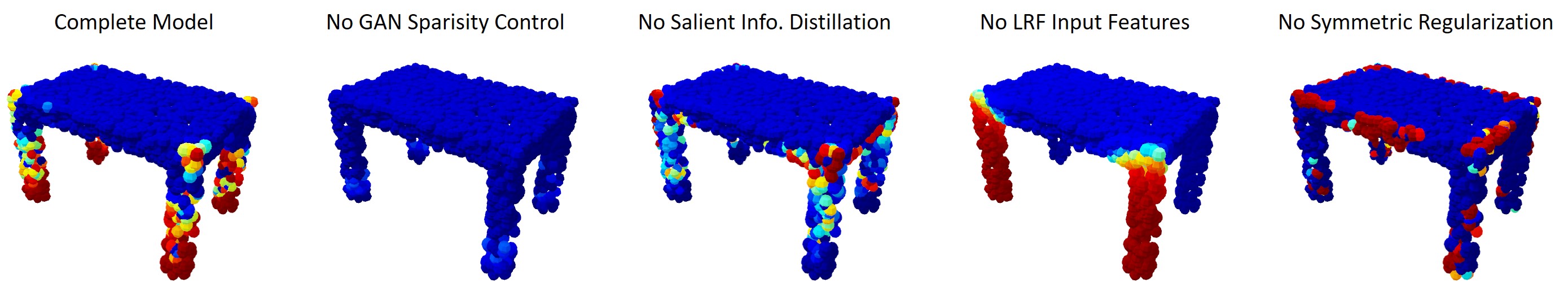}
    \caption{\textbf{Visualizations of ablation study on a ShapeNet table.} Colors indicate keypoint probabilities (red means high and blue means low). We see that without GAN sparsity control, our model fails to give meaningful keypoints.}
    \label{fig:ablation}
\end{figure*}

\subsection{Keypoints for Real-World Registration}
\paragraph{Dataset}
3DMatch dataset~\cite{zeng20173dmatch} is an indoor registration benchmark. The test set contains 8 scenes with partially overlapped point cloud fragments and their corresponding transformation matrices. ETH dataset~\cite{pomerleau2012challenging} is another outdoor registration benchmark, whose test set contains 4 scenes with overlapped fragments. Our keypoint detector is trained on ShapeNet dataset and then directly applied to 3DMatch and ETH. We down-sample 3DMatch and ETH point clouds using a voxel grid filter of size 0.03m and 0.02m, respectively.

\paragraph{Metric}
Geometric registration usually consists of two stages: keypoint detection and descriptor extraction. In order to compare the performance of different keypoint detectors, we leverage two state-of-the-art descriptors: PerfectMatch~\cite{gojcic2019perfect} and D3Feat~\cite{bai2020d3feat}, and combine them with our detector. As a baseline, random sampling, traditional detectors (i.e., ISS, SIFT-3D) or task-specific learning based detectors (i.e., D3Feat) are also evaluated. For each fragment, different number of points are given (i.e., 2500, 1000, 500, 250, 100) as budgets. We use NMS on keypoint score for D3Feat and UKPGAN, and random sampling is leveraged for traditional detectors to fulfill the budget requirement. We follow D3Feat to use Feature Matching Recall, Registration Recall and Inlier Ratio for evaluation. Feature Matching Recall is the percentage of successful alignment whose inlier ratio is above some threshold (i.e., $\tau_2 = 5\%$), which measures the matching quality of pairwise registration. A point-pair alignment is deemed successful if their distance is within some threshold (i.e., $\tau_1 = 0.1m$). Registration Recall is the percentage of successful alignment whose transformation error is below some threshold (i.e., RMSE $< 0.2m$), which better reflects the final performance. We use RANSAC with 50,000 max iterations to estimate the transformation matrices.


\paragraph{Evaluation and Results}

\begin{table*}[ht!]
\begin{center}
\resizebox{0.9\linewidth}{!}{
\begin{tabular}{c|c|ccccc|ccccc|ccccc}
\toprule[1pt]
 \multicolumn{2}{c|}{} & \multicolumn{5}{c|}{Feature Matching Recall (\%)} & \multicolumn{5}{c|}{Registration Recall (\%)} & \multicolumn{5}{c}{Inlier Ratio (\%)}\\
\hline
Detector & Descriptor & 2500 & 1000 & 500 & 250 & 100 & 2500 & 1000 & 500 & 250 & 100 & 2500 & 1000 & 500 & 250 & 100 \\
\hline
ISS~\cite{zhong2009intrinsic} & PerfectMatch & 90.3 & \textbf{87.9} & 82.7 & 71.9 & 51.7 & 75.5 & 68.6 & 55.4 & 37.9 & 14.0 & 28.1 & 22.6 & 18.7 & 15.7 & 12.9 \\
SIFT~\cite{rister2017volumetric} & PerfectMatch & 90.3 & 87.7 & 82.5 & 74.5 & 52.7 & \textbf{77.4} & 68.1 & 56.4 & 35.8 & 11.7 & 28.0 & 22.6 & 18.5 & 15.1 & 12.4 \\
Random & PerfectMatch & \textbf{90.4} & 86.8 & 82.3 & 71.2 & 53.5 & 76.8 & 68.9 & 54.8 & 36.2 & 16.1 & 28.2 & 22.8 & 18.5 & 15.1 & 12.3 \\
Ours & PerfectMatch & 90.1 & 87.8 & \textbf{85.6} & \textbf{83.1} &\textbf{ 74.2} & 76.1 & \textbf{72.5} & \textbf{70.0} & \textbf{63.6} & \textbf{37.6} & \textbf{28.5} & \textbf{25.4} & \textbf{25.7} & \textbf{24.5} & \textbf{18.8}  \\
\hline
ISS~\cite{zhong2009intrinsic} & D3Feat & 95.2 & 94.4 & 93.4 & 90.1 & 81.0 & 83.5 & 79.2 & 76.0 & 64.3 & 37.2 & 38.2 & 33.5 & 28.8 & 23.9 & 17.4\\
SIFT~\cite{rister2017volumetric} & D3Feat & 94.9 & 94.0 & 93.0 & 91.2 & 81.3 & 84.0 & 79.9 & 76.1 & 60.9 & 38.6 & 38.4 & 33.6 & 28.8 & 23.3 & 17.4 \\
Random & D3Feat & 95.1 & \textbf{94.5} & 92.8 & 90.0 & 81.2 & 83.0 & 80.0 & 77.0 & 65.5 & 38.8 & 38.6 & 33.6 & 28.9 & 23.6 & 17.3 \\
D3Feat~\cite{bai2020d3feat} & D3Feat & \textbf{95.5} & \textbf{94.5} & \textbf{94.1} & \textbf{93.1} & \textbf{90.6} & \textbf{84.3} & \textbf{83.6} & \textbf{82.5} & \textbf{78.1} & \textbf{67.2} & \textbf{40.5} & \textbf{42.6} & \textbf{44.0} & \textbf{44.7} & \textbf{45.6} \\
Ours & D3Feat & 94.7 & 94.2 & 93.5 & 92.6 & 85.9 & 82.8 & 81.4 & 77.1 & 69.7 & 47.4 & 38.8 & 35.5 & 34.0 & 33.1& 27.7 \\
\hline
\end{tabular}}
\end{center}   
 \caption{\textbf{Registration result on 3DMatch.} We evaluate on two state-of-the-art descriptors, combined with different keypoint detectors. }
\label{tab:realworld}
\end{table*}

\begin{table*}[ht!]
\begin{center}
\resizebox{0.9\linewidth}{!}{
\begin{tabular}{c|c|ccccc|ccccc|ccccc}
\toprule[1pt]
 \multicolumn{2}{c|}{} & \multicolumn{5}{c|}{Feature Matching Recall (\%)} & \multicolumn{5}{c|}{Registration Recall (\%)} & \multicolumn{5}{c}{Inlier Ratio (\%)}\\
\hline
Detector & Descriptor & 2500 & 1000 & 500 & 250 & 100 & 2500 & 1000 & 500 & 250 & 100 & 2500 & 1000 & 500 & 250 & 100 \\
\hline
ISS~\cite{zhong2009intrinsic} & PerfectMatch & 59.1 & 41.3 & 23.1 & 11.3 & 6.3 & 48.8 & 28.1 & 12.8 & 5.2 & 1.3 & 11.4 & 9.0 & 8.0 & 6.8 & 6.8 \\
SIFT~\cite{rister2017volumetric} & PerfectMatch & 58.5 & 39.4 & 23.9 & 10.8 & 6.2 & 45.5 & 26.9 & 12.9 & 6.0 & 0.9 & 11.3 & 9.0 & 7.5 & 6.8 & 6.5 \\
Random & PerfectMatch & 60.8 & 39.7 & 22.2 & 13.7 & 4.4 & 50.1 & 30.7 & 16.6 & 4.3 & 0.4 & 11.3 & 9.1 & 7.6 & 6.8 & 6.4 \\
Ours & PerfectMatch & \textbf{68.1} & \textbf{62.4} & \textbf{53.6} & \textbf{44.8} & \textbf{29.6} & \textbf{58.2} & \textbf{45.5} & \textbf{32.3} & \textbf{19.1} & \textbf{6.1} & \textbf{18.7} & \textbf{16.2} & \textbf{14.2} & \textbf{11.8} & \textbf{10.0}  \\
\hline
ISS~\cite{zhong2009intrinsic} & D3Feat & 37.9 & 24.4 & 16.3 & 10.8 & 6.2 & 25.6 & 18.1 & 8.9 & 4.7 & 1.7 & 8.8 & 7.7 & 7.2 & 6.6 & 7.5 \\
SIFT~\cite{rister2017volumetric} & D3Feat & 36.8 & 24.6 & 14.9 & 10.2 & 5.5 & 28.4 & 16.7 & 9.0 & 3.0 & 1.1 & 8.7 & 7.7 & 7.0 & 7.2 & 6.7 \\
Random & D3Feat & 27.7 & 16.7 & 7.7 & 3.6 & 2.1 & 20.4 & 11.0 & 7.0 & 1.5 & 1.5 & 8.1 & 6.7 & 6.5 & 6.3 & 6.3 \\
D3Feat~\cite{bai2020d3feat} & D3Feat & \textbf{48.5} & \textbf{54.5} & \textbf{57.0} & \textbf{57.3} & \textbf{49.9} & \textbf{29.2} & \textbf{28.7} & \textbf{29.5} & \textbf{22.8} & \textbf{11.2} & 10.9 & \textbf{12.0} & \textbf{13.0} & \textbf{13.5} & \textbf{13.9} \\
Ours & D3Feat & 47.5 & 43.1 & 37.4 & 33.0 & 21.5 & 28.3 & 22.0 & 14.2 & 10.9 & 3.9 & \textbf{12.4} & 11.6 & 10.9 & 9.9 & 9.2 \\
\hline
\end{tabular}}
\end{center}   
 \caption{\textbf{Registration result on ETH.} We evaluate on two state-of-the-art descriptors, combined with different keypoint detectors. }
\label{tab:realworld2}
\end{table*}
Results are shown in Table~\ref{tab:realworld} and \ref{tab:realworld2}. The PerfectMatch and D3Feat descriptors are based on the pretrained model released by the authors. For PerfectMatch descriptor, our keypoint detector outperforms other detectors by a large margin, especially when the number of keypoints is small. For D3Feat descriptor, though D3Feat detector performs the best, the detector is trained together with the descriptor on real-world training data, while our keypoint detector is trained on synthetic ShapeNet models only. Besides, our method also outperforms other traditional keypoint detectors by a large margin. Our model could generalize to real-world data and may improve registration results. Qualitative results are given in Figure~\ref{fig:regis}.


\subsection{Repeatability under Arbitrary Rotations}
\label{sec:rot}

A good keypoint detector should be invariant to rotations, since orientations are often unknown. Therefore, rotation repeatability is an important metric to measure the quality of a keypoint detector. We evaluate on the test split of KeypointNet dataset, averaged over airplane, chair and table.


We follow the relative repeatability metric proposed in USIP as the evaluation metric. Given two point clouds of the same object, a keypoint in the first point cloud is considered repeatable if its distance to the nearest keypoint in the second point cloud is less than 0.1, under  ground-truth transformations. We report the percentage of repeatable keypoints when different number of keypoints are detected.



We compare UKPGAN with USIP~\cite{li2019usip}, D3Feat~\cite{bai2020d3feat}, Harris-3D~\cite{sipiran2011harris}, ISS~\cite{zhong2009intrinsic} and SIFT-3D~\cite{rister2017volumetric}. We generate 4, 8, 16, 32, 64, 128, 256, 512 most salient keypoints and calculate the relative repeatability respectively.
The relative repeatability under arbitrary rotations on KeypointNet dataset is shown in Figure~\ref{fig:rot_rep}. Thanks to the local reference frame (LRF) extracted in our method, we achieve much higher keypoint repeatability than all previous methods. Even only four keypoints are detected, we achieve nearly 100\% repeatability.

\subsection{Ablation Study}
\label{sec:ablation}
In this section, we validate our design choices by conducting several ablation studies. Evaluation results are done on KeypointNet test split. Both IoU and rotation repeatability are evaluated. IoU is reported by NMS under threshold 0.1 and rotation repeatability is reported with 4 most salient keypoints. Quantitative and qualitative results are shown in Table~\ref{tab:ablation} and Figure~\ref{fig:ablation}.


\paragraph{GAN-based Keypoint Sparsity Control.}
GAN allows learning keypoint distributions with easily controllable parameters.
We experienced with L1 norm and found that it fails to output a meaningful keypoint distribution by tuning the coefficient of norm loss, as shown in Figure~\ref{fig:ablation}. 

\paragraph{Salient Information Distillation.}
Salient information Distillation is another important module for our model. We compare our complete model with a baseline that implements a simple averaging instead of max-pooling. It shows that with no salient information distillation, salient parts of models are not detected.

\paragraph{Local Rotation Invariant Descriptors.}
Local rotation invariant descriptors play an important role in maintaining repeatability under arbitrary rotations. If we replace it with raw $XYZ$ features, both IoU and rotation repeatability drop.


\paragraph{Symmetric Regularization.}
In Section~\ref{sec:sym}, we integrate a symmetric invariance prior into our model, which is helpful since the extracted descriptors are only rotation invariant rather than symmetric invariant. If we remove symmetric regularization, we see that detected keypoints are not symmetric anymore in Figure~\ref{fig:ablation}.
, with a drop in both IoU and rotation repeatability.

\begin{table}[t]
\begin{center}
\resizebox{\linewidth}{!}{
\begin{tabular}{l|ccc|ccc}
\toprule[1pt]
\multirow{2}*{}  & \multicolumn{3}{c|}{IoU (\%)} & \multicolumn{3}{c}{Rotation Rep. (\%)}\\
\cline{2-7}
~ & airplane & chair & table & airplane & chair & table \\
\hline
Ours & \textbf{68.8} & \textbf{36.2} & \textbf{34.7} & 98.3 & 88.3 & 90.6\\
\hline
Ours w/o GAN Sparsity & 36.3 & 27.2 & 23.1 & \textbf{99.8} & \textbf{95.9} & 99.6 \\
Ours w/o Salient Info. Distill. & 51.8 & 33.3 & 19.0 & 94.1 & 87.5 & \textbf{99.9} \\
Ours w/o LRF Feat. & 22.4 & 16.0 & 21.2 & 15.4 & 4.9 & 0.7 \\
Ours w/o Symmetric Reg. & 54.9 & 20.0 & 22.3 &85.1 & 77.0 & 73.0 \\
\hline
\end{tabular}}
\end{center}   
\caption{\textbf{Results of ablation studies.}}
\label{tab:ablation}
\end{table}

\section{Conclusion}
In this work, we proposed an adversarial keypoint detector with adversarial sparsity loss which could detect meaningful points in an unsupervised way. The key contributions of our method are GAN-based sparsity control and salient information distillation modules. Experiments show that our UKPGAN detector can produce stable points on rigid and non-rigid objects. Moreover, our method also generalizes well to real scenarios.

\section{Acknowledgements}
This work was supported by the National Key Research and Development Project of China (No. 2021ZD0110700), the National Natural Science Foundation of China under Grant 51975350, Shanghai Municipal Science and Technology Major Project (2021SHZDZX0102), Shanghai Qi Zhi Institute, and SHEITC (2018-RGZN-02046). This work was also supported by the  Shanghai AI development project (2020-RGZN-02006) and ``cross research fund for translational medicine'' of Shanghai Jiao Tong University (zh2018qnb17, zh2018qna37, YG2022ZD018).

{\small
\bibliographystyle{ieee_fullname}
\bibliography{egbib}
}

\appendix
\addcontentsline{toc}{section}{Appendices}







\section*{Supplementary}

\section{More Analysis on Keypoint Controllability with Beta Distribution}
\begin{figure}[ht]
    \centering
    \includegraphics[width=0.8\linewidth]{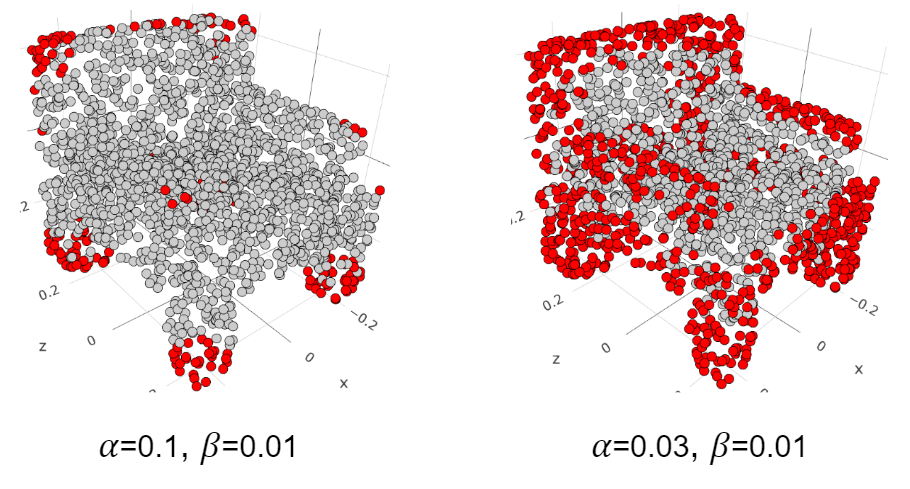}
    \captionof{figure}{We can easily control the number of keypoints with Beta distribution parameters. Keypoints are shown in red with $p>0.5$.}
    \label{fig:control}
\end{figure}

In Beta distributions, we have $\alpha$ and $\beta$ controlling the keypoint probability accumulation around 0 and 1, respectively. This  controllability is  illustrated  in Figure~\ref{fig:control}. We see that our model automatically identify those semantic keypoints with certain distribution requirements. By introducing Beta distribution and GAN sparsity control, our model is able to detect corner points when budget is tight, and detect less salient points (e.g., edge points) when more budget is given. In contrast, previous methods on keypoint detection do not give much control on the number of keypoints.

\section{Precise Control of Number of Keypoints}
Our model outputs a keypoint distribution, which lies in $[0, 1]$.
In order for a precise control of specific number of keypoints, we could use iterative Non-Maximum-Suppression (NMS) with radius $r$. Specifically, suppose we want exactly $K$ keypoints, at each iteration, we pick the point with largest $\Phi$ and
then invalidate all geodesic neighbors within radius $r$, this iteration is repeated until we have K points.

\section{Results on SMPL Models with NMS Applied}
\begin{figure}
      \centering
  \includegraphics[width=0.9\linewidth]{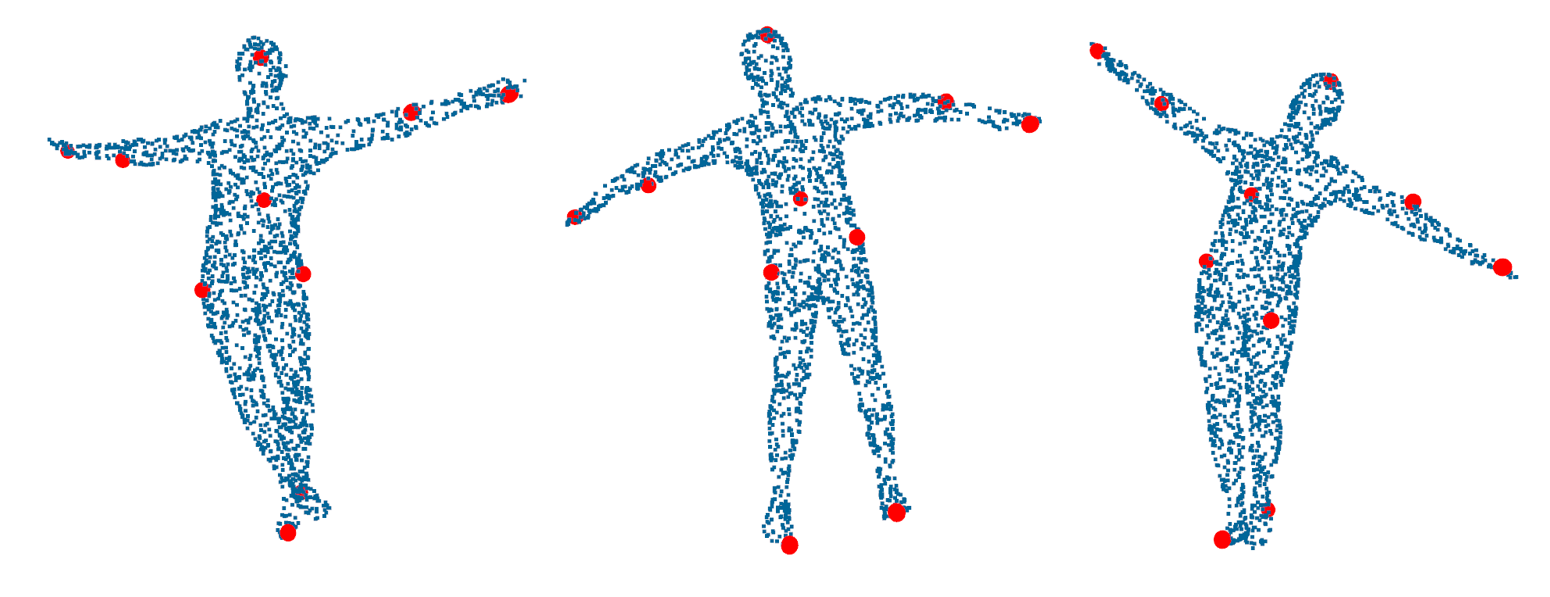}
  \captionof{figure}{Results on SMPL models. NMS is applied to ensure 10 keypoints.}
  \label{fig:smplvis_nms}
\end{figure}

We evaluate all the methods on SMPL models by applying NMS with fixed number of 10, 20 and 40 keypoints. Qualitative comparison is given in Figure~\ref{fig:smplvis_nms} and quantitative results are listed in Table~\ref{tab:smplnms}.

\begin{table}[ht]
\centering
\resizebox{0.9\linewidth}{!}{
\begin{tabular}{l|c|c|c|c|c|c}
    \toprule
     \multirow{2}*{} & \multicolumn{3}{c|}{IoU (\%) $\uparrow$ } & \multicolumn{3}{c}{Cons. ($\times 1e^{-2}$) $\downarrow$}\\
    \cmidrule{2-7}
     & 10 & 20 & 40 & 10 & 20 & 40 \\
    \midrule
    USIP & 2.0 & 2.4 & 2.1 & 13.9 & 12.5 & 12.7 \\
    D3Feat & 1.6 & 1.8 & 1.9 & 15.8 & 14.1 & 11.6  \\
    HARRIS-3D & 0.4 & 0.5 & 0.9 & 21.5 & 15.1 & 10.4 \\
    ISS & 1.0 & 0.9 & 1.2 & 20.8 & 14.8 & 10.3 \\
    SIFT-3D & 0.1 & 0.4 & 1.0 & 21.9 & 14.9 & 10.4\\
    \midrule
    Ours & \textbf{7.2} & \textbf{7.5} & \textbf{9.2}& \textbf{9.2} & \textbf{7.9} & \textbf{5.9}\\
    \bottomrule
\end{tabular}}
\caption{IoU (\%) and Consistency Loss ($\times 1e^{-2}$) results for SMPL dataset, given the budget of 10, 20, 40 keypoints.}
\label{tab:smplnms}
\end{table}

\section{Results on More ShapeNet Categories}
\begin{figure}
      \centering
  \includegraphics[width=0.9\linewidth]{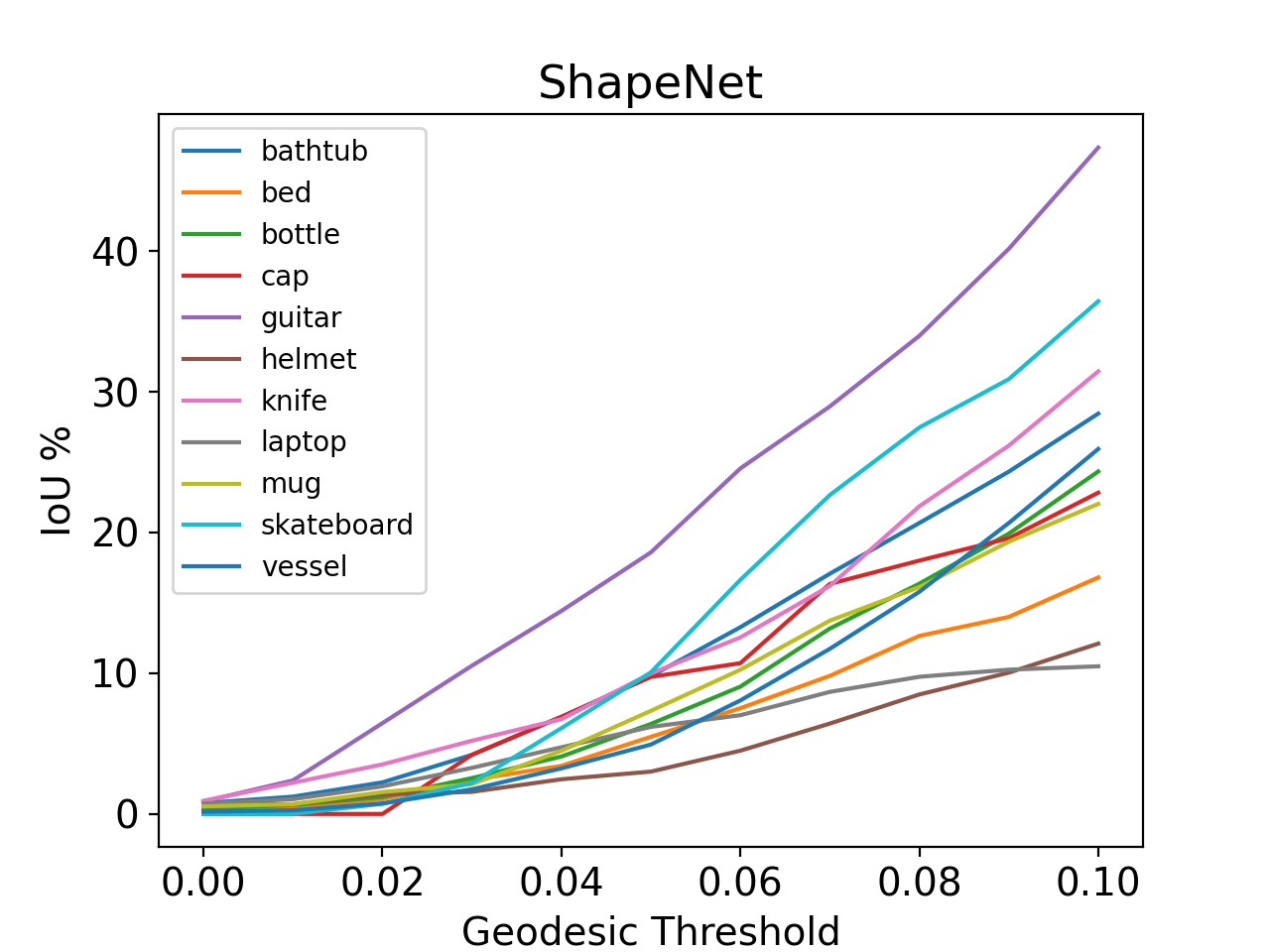}
  \captionof{figure}{Quantitative results with a universal model on ShapeNet models. NMS with radius 0.1 is applied.}
  \label{fig:shapenetmore}
\end{figure}

\begin{figure}
      \centering
  \includegraphics[width=0.9\linewidth]{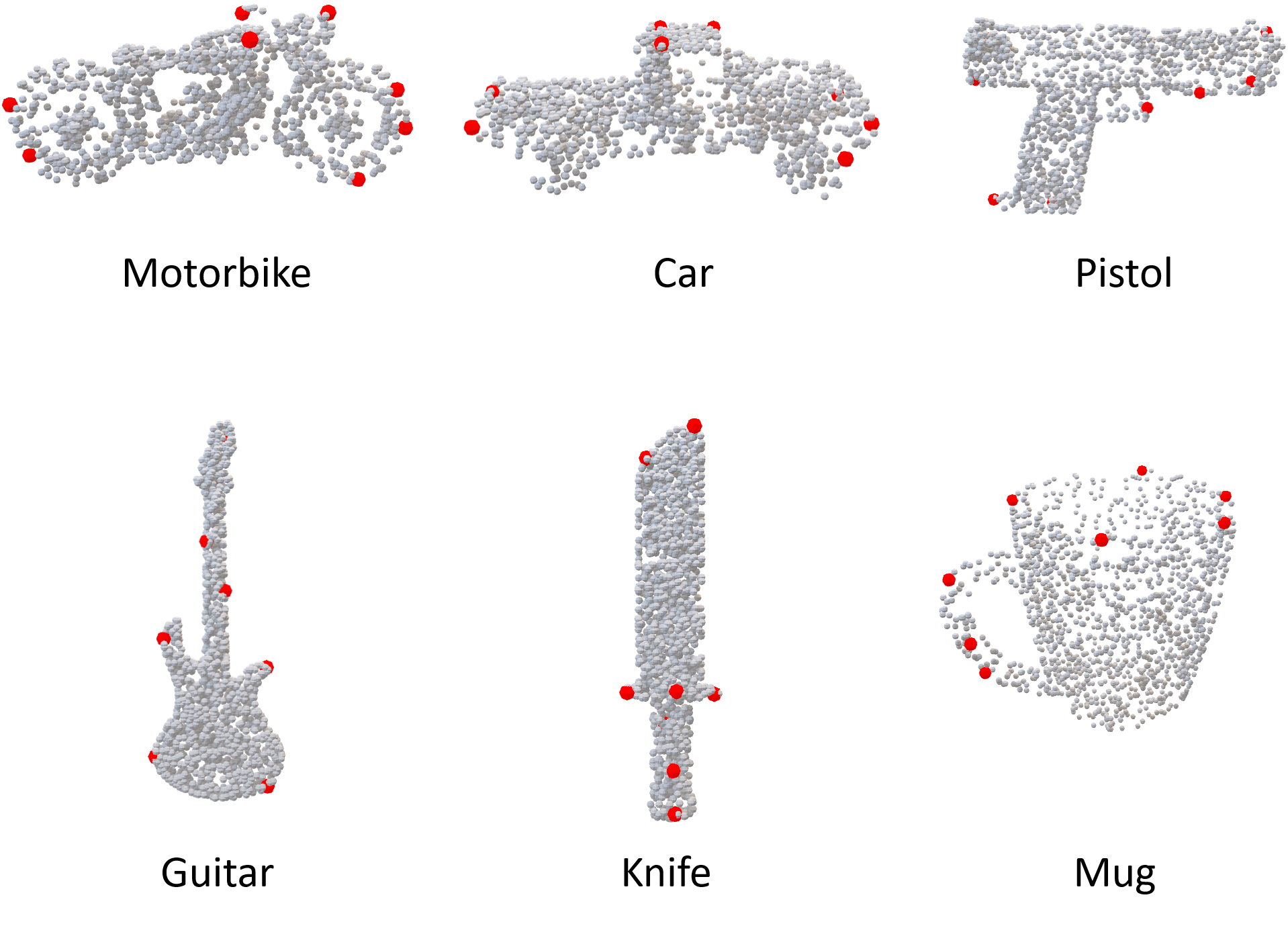}
  \captionof{figure}{Qualitative keypoint detection results on more ShapeNet categories. NMS with radius 0.1 is applied.}
  \label{fig:shapenetmore_vis}
\end{figure}

We also evaluate a universal model that is trained on a large collection of ShapeNet models, including 11 categories. Quantitative results are given in Figure~\ref{fig:shapenetmore}.
Qualitative results are illustrated in Figure~\ref{fig:shapenetmore_vis}.

\section{Extension on Unsupervised 2D Binary Image Analysis}

Our method is not only restricted to 3D shape analysis, but 2D binary images. In this experiment, we evaluate our method on MNIST, by viewing each pixel as a 2D point. The encoder is replaced with 2D convolution without local reference frame. Digits from all classes are trained jointly.

\begin{figure*}[h!]
\centering
\begin{minipage}{.45\textwidth}
  \centering
  \includegraphics[width=0.9\linewidth]{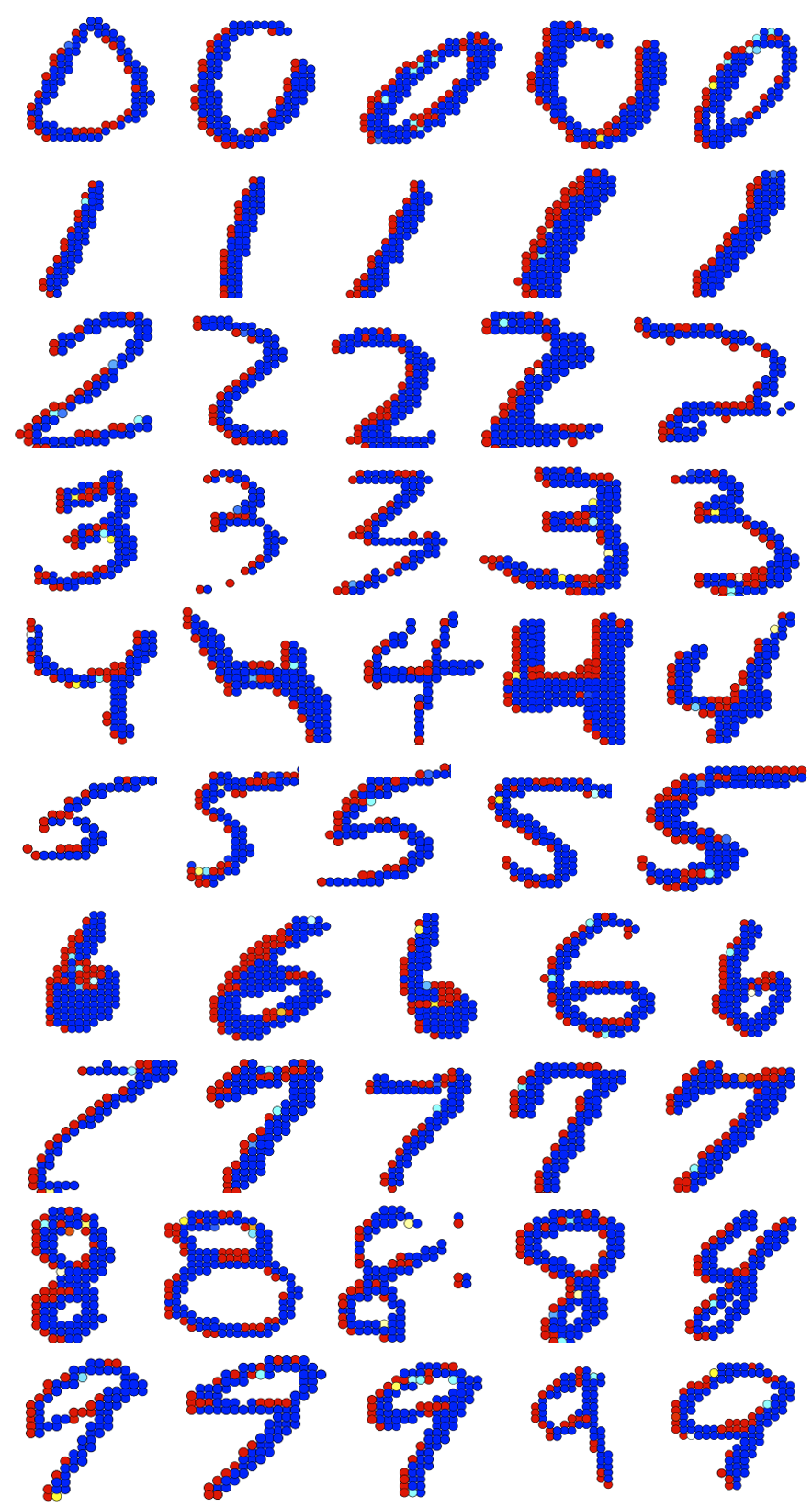}
  \captionof{figure}{\textbf{Keypoint probability heat-map on MNIST.} Red indicates high probability.}
  \label{fig:mnist_heat}
\end{minipage}%
\hfill
\begin{minipage}{.45\textwidth}
  \centering
  \includegraphics[width=0.9\linewidth]{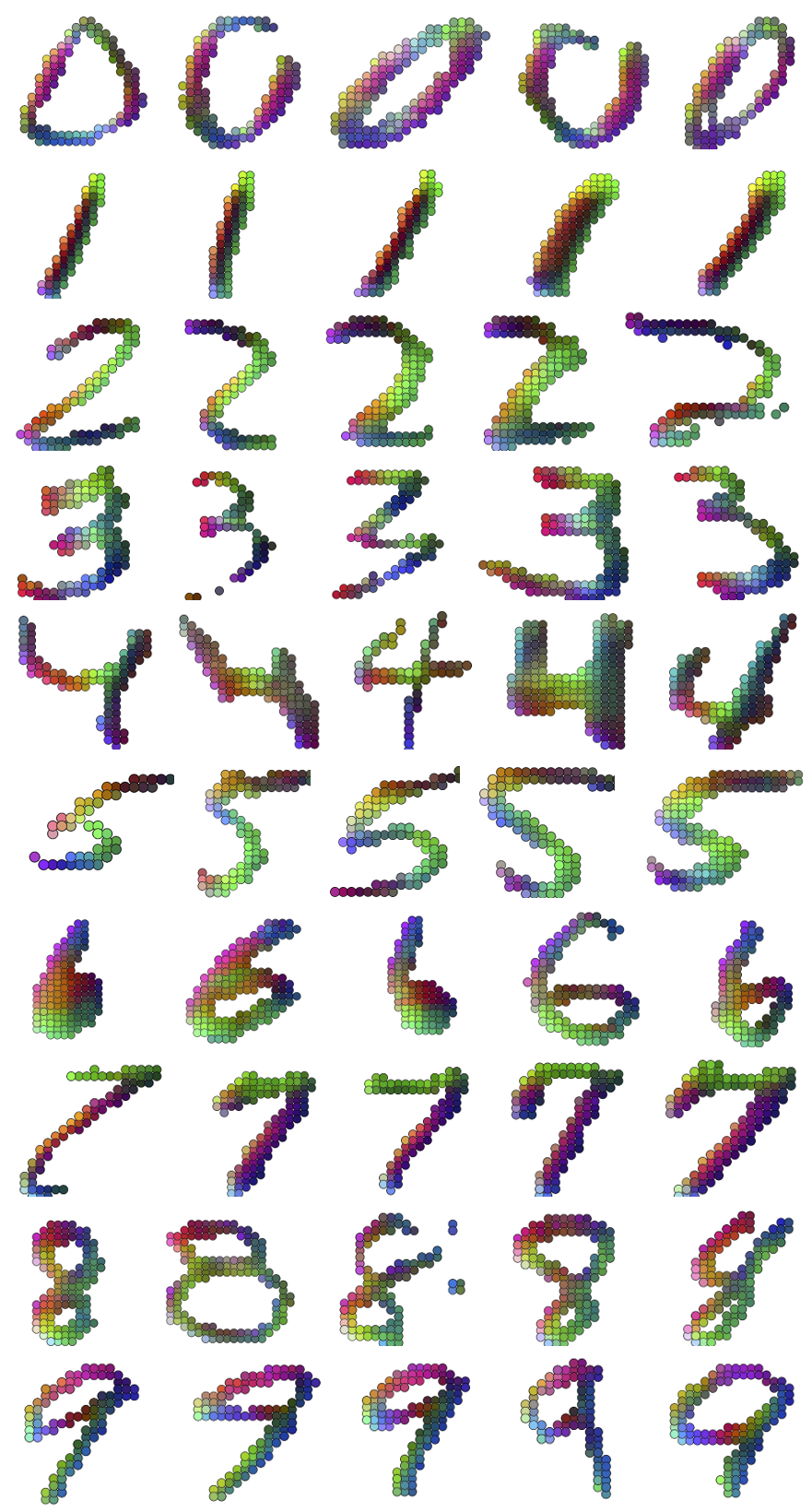}
  \captionof{figure}{\textbf{Dense embeddings generated on MNIST, which are consistent across digits within the same class.} Best viewed in color.  }
  \label{fig:mnist_emb}
\end{minipage}
\end{figure*}

The results on unsupervised keypoint detection is shown in Figure~\ref{fig:mnist_heat}. It can be shown that our algorithm captures important keypoint skeletons in MNIST digits, and they are consistent within each class. In the meantime, unsupervised dense embeddings are also predicted for each pixel. Quite interestingly, the generated embeddings (Figure~\ref{fig:mnist_emb}) are consistent within each class, without acquiring any class label at training time.

\section{More Visualizations on Detected Keypoints under Arbitrary Rotations}
\begin{figure*}[htb]
    \centering
    \includegraphics[width=0.95\linewidth]{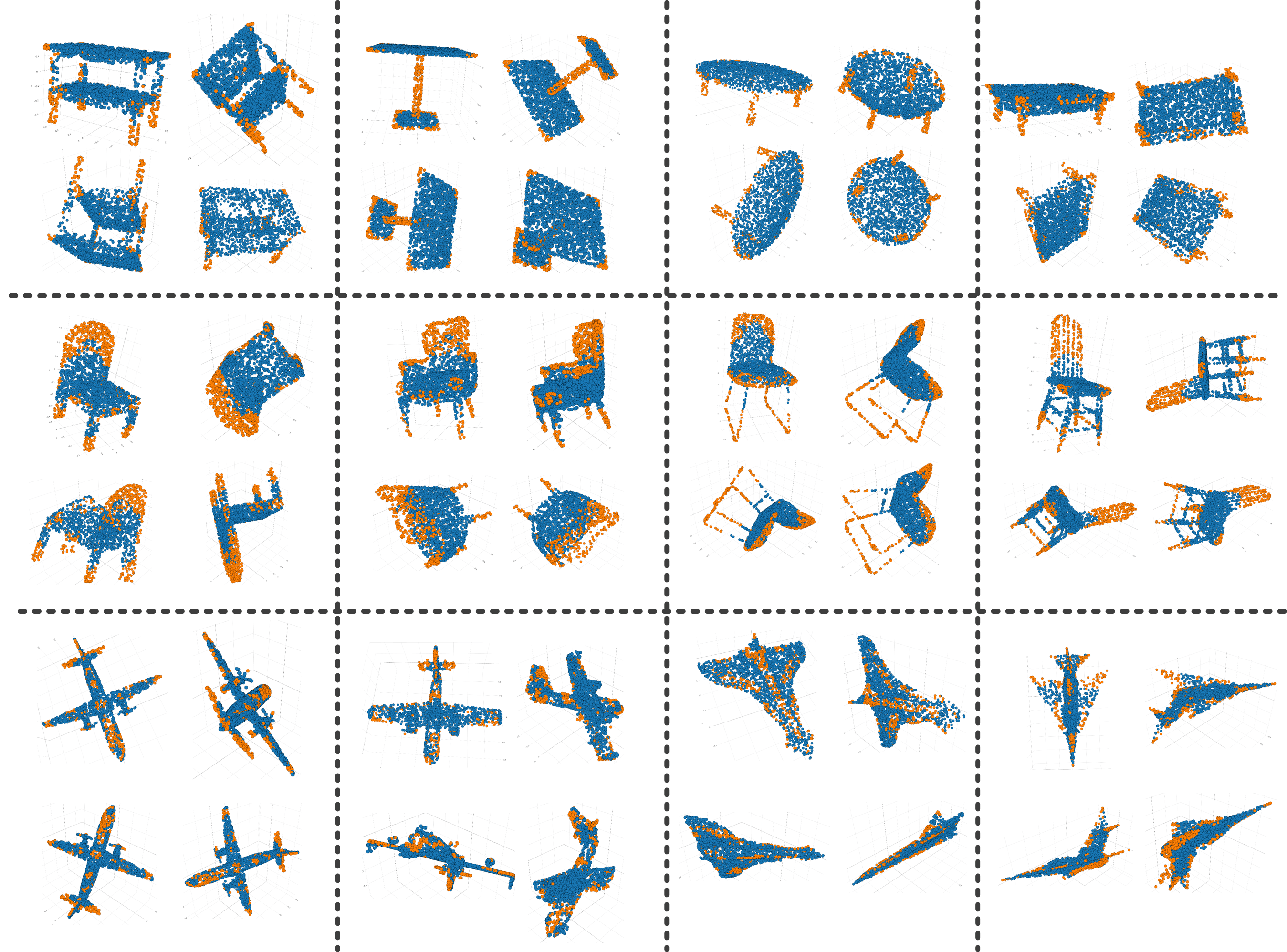}
    \captionof{figure}{More visualization results. Each model is rotated four times. Point clouds are shown in blue while keypoints are shown in orange.}
    \label{fig:rotation4}
\end{figure*}

We plot more visualization results in Figure~\ref{fig:rotation4}, where each model is rotated four times. Keypoints are filtered by $p>0.5$ with no NMS is applied. We see that UKPGAN does a pretty good job in maintaining rotation repeatability.

\section{More Qualitative Results on Real-World Keypoint Detection}
Here, we plot more keypoint detection results on real-world scenarios, under both indoor (3DMatch) and outdoor (ETH) settings. Notice that our model is trained on synthetic models only, while it generalizes to real-world scenarios well. We can see that our method detect salient corner points on both indoor and outdoor datasets, thus boosting the performance of geometric registration.
\begin{figure}[ht]
    \centering
    \includegraphics[width=0.95\linewidth]{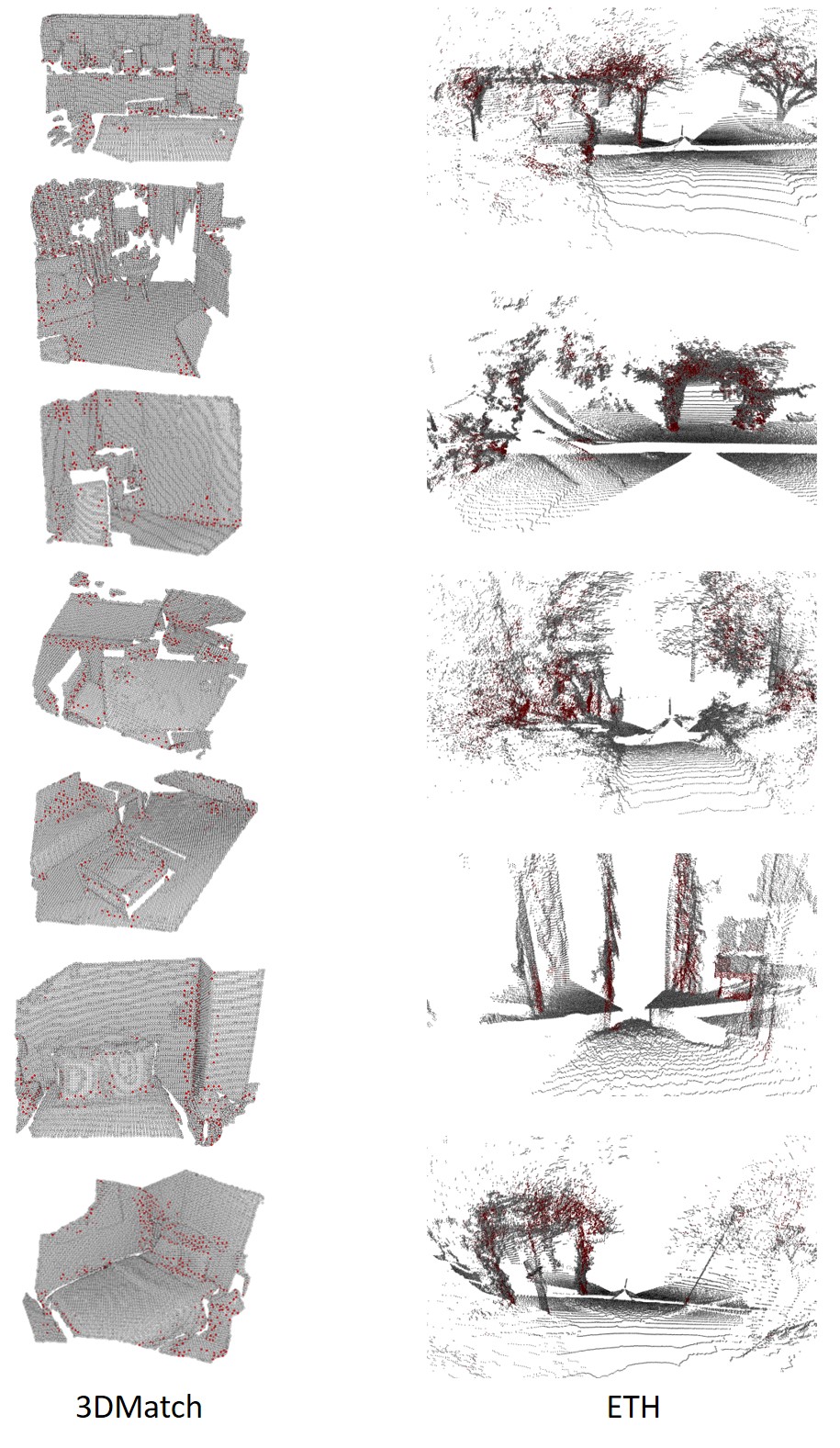}
    \captionof{figure}{Keypoint detection results on real-world scenes. Left: indoor 3DMatch dataset. Right: outdoor ETH dataset.}
    \label{fig:reg}
\end{figure}

\end{document}